\documentclass[french]{article-hermes}
\usepackage{textcomp}
\usepackage{graphicx}

\makeatletter

\usepackage{babel}

\makeatother

\usepackage{babel}

\begin{document}
\submitted{ MARAMI 2013}{1}

\title[détection de communautés recouvrantes]
{Optimisation dans la détection de communautés recouvrantes et équilibre
de Nash}
\subtitle{}

\author{Michel Crampes, Michel Plantié, Marie Lopez}

\address{Laboratoire LGI2P, Ecole des Mines, Parc Georges Besse, 30035 Nîmes}

\resume{
La détection de communautés dans les graphes a fait l'objet de nombreux
algorithmes. Les méthodes récentes cherchent à optimiser une fonction
de modularité qui témoigne d'un maximum de relations à l'intérieur
des communautés et d'un minimum de relations inter-communautés.
D'abord appliqués aux graphes mono-parti non orientés, ces algorithmes
étendent leur champ d'application aux graphes multi-partis et orientés.
Cependant étant donné le caractère NP-complet du problème, ces algorithmes
sont des heuristiques qui ne garantissent pas l'obtention d'un optimum
global voire même local. Dans cet article nous introduisons
un algorithme qui, à partir d'une solution approchée obtenue à l'aide
d'un algorithme de détection efficace, la modifie pour atteindre un
optimum local basé sur une fonction de réaffectation. La fonction
de réaffectation est une 'fonction potentiel' et en conséquence l'optimum
obtenu est un équilibre de Nash. Nous complétons notre méthode par
une fonction de recouvrement qui permet de disposer de manière concomittante
des deux modes de détection. Quelques expériences montre l'intérêt
de l'approche.
}
\abstract{
Community detection in graphs has been the subject of many
algorithms. Recent methods want to optimize a 
modularity function which shows a maximum of relationships within
communities and found a minimum of inter-community relations.
these algorithms are applied to unipartite, multipartite and directed graphs.
However, given the NP-completeness of the problem, these algorithms
are heuristics that do not guarantee an optimum.
In this paper we introduce an algorithm which, based on an approximate solution obtained through
a efficient detection algorithm, modifie it to achieve a
local optimum based on a  function. this reassignment function
is a 'potential function' and therefore the computed optimum
is a Nash equilibrium. We supplement our method with an overlap function
that allows to have simultaneously the two detection modes. Several experiments show the interest
of our approach.
}
\motscles{Recouvrement de communautés, détection de communautés, Equilibre de Nash}
\maketitlepage

\section{Introduction}
Avec le développement d'Internet et l'importance prise par les réseaux
sociaux, la recherche de communautés à partir de graphes fait l'objet
de nombreux travaux. Beaucoup d'algorithmes aux approches très variées
ont été proposés. Les plus importants sont rapportés dans \cite{Papadopoulos2011,Yang2010,Porter2009})
et de manière plus détaillée dans \cite{Fortunato2009}. En grande
majorité ils portent sur la détection de communautés partitionnées
(ensembles disjoints de sommets). Dans la réalité les individus appartiennent
à plusieurs communautés. Le recouvrement est plutôt la règle. Quelques
travaux récents se sont intéressés à la détection de communautés recouvrantes,
pour l'essentiel dans les graphes unipartis. Dans un article précédent
nous nous étions intéressés à la recherche de communautés recouvrantes
dans les graphes bipartis en utilisant les treillis de Galois \cite{Crampes2012}.
Dans un article plus récent \cite{Crampes2013} nous proposons une
méthode différente. Elle utilise tout algorithme de partitionnement
de graphes unipartis pour traiter à la fois les graphes unipartis,
les graphes bipartis et les graphes orientés et produire des communautés
partitionnées et recouvrantes. Concrètement nous utilisons la méthode
de Louvain \cite{Blondel2008}. Appliquée à des benchmarks bien connus
nous obtenons des résultats au moins aussi bons et souvent meilleurs
que ceux obtenus par d'autres auteurs. En particulier nous présentons
à la fois des communautés partitionnées et recouvrantes. Dans d'autres
domaines comme par exemple la tractographie du cerveau, les résultats
sont pertinents au regard d'une analyse sémantique manuelle. Cependant,
malgré son efficacité, la méthode basée sur l'algorithme de Louvain
présente une limite que nous souhaitons lever dans le présent article.
Comme toute heuristique, elle donne des communautées partitionnées
pour lesquelles la répartition des sommets dans le graphe ne garantit
pas un optimum global pour la modularité. Le caractère non optimal
est en particulier clairement montré par les recouvrements. Pour nous
approcher de cet optimum nous introduisons une fonction de réaffectation
qui permet de prendre en compte le souhait d'un sommet pour être changé
de communauté. Nous complétons cette fonction par une condition de
stabilité et nous montrons que lorsque cette stabilité est atteinte
nous obtenons un équilibre de Nash. Les résultats sont nettement améliorés.
Le fait de faire appel à l'équilibre de Nash n'est pas seulement un
moyen d'obtenir un optimum local dans des temps acceptables. C'est
aussi un source d'interprétation sémantique de l'affectation et de
la réaffectation des agents dans les communautés. Notre article se
conclue sur l'ouverture de pistes de réflexion en ce sens.

\section{État de l'art }
Les nombreux travaux de recherche se sont focalisés
dans un premier temps sur la détection de communautés partitionnées
dans des graphes unipartis. Fortunato dans \cite{Fortunato2009}
en a fait une liste exhaustive. La plupart des algorithmes récents proposent de maximiser une mesure
appelée modularité. La notion de modularité pour optimiser l'organisation
d'un graphe en communautés introduite par Newman \cite{Newman2006}
a été largement appliquée d'abord aux graphes unipartis. Puis quelques
auteurs l'ont étendue aux graphes bipartis dans un premier temps en
adaptant sa formulation\cite{Murata2010,Suzuki2009}. Cependant à
partir de Barber \cite{Barber2007} une expression de la modularité
pour les graphes bipartis est directement dérivée de Newman et reprise
par divers auteurs pour appliquer des méthodes classiques telles que
le recuit simulé \cite{Guimera2007}, le clustering spectral \cite{Barber2007},
les algorithmes génétiques \cite{Nicosia2009}, la transmission de
labels \cite{LiuXin2010}, ou encore l'analyse spectrale dichotomique
\cite{Leicht2007}. De leur coté les stratégies de détection de communautés recouvrantes
à partir de graphes unipartis éventuellement pondérés étendent bien
souvent les méthodes de partitionnement. \cite{Palla2005a} et utilisent
le système de percolation de k-cliques. \cite{Davis2008} font appel
à la marche aléatoire dans un graphe. \cite{Gregory2009} utilise
les algorithmes de propagation de labels. Certains auteurs proposent
des méthodes spécifiques. \cite{Wu2012a} cherchent les recouvrements
entre communautés partitionnées. \cite{Reichardt2006a} combinent
le modèle d'interaction de spins de Pott et le recuit simulé. \cite{Lancichinetti2009}
optimisent un fonction statistique locale. \cite{Evans2009} traitent
le problème dual de partitionner les liaisons pondérées. Il est plus difficile de trouver dans la littérature des travaux qui
portent à la fois sur les graphes bipartis et le recouvrement des
communautés. On retrouve l'extension de méthodes telle la recherche
de bicliques recouvrantes \cite{SuneLehmannMartinSchwartzLarsKaiHansen2008}.
D'autres méthodes originales font appel aux résultats connus dans
les treillis de Galois \cite{Crampes2012} et \cite{Roth2008}, mais
la représentation et les algorithmes sont complexes. Dans un article
récent \cite{Crampes2013} nous montrons qu'il est possible d'unifier
les graphes unipartis, bipartis, et orientés pour produire des communautés
à la fois partitionnées et recouvrantes, le recouvrement utilisant
une fonction simple d'appartenance simultanée. Nous mettons en oeuvre
l'algorithme de Louvain \cite{Blondel2008} qui fonctionne en agrègeant
itérativement les sommets du graphe afin d'augmenter la modularité
au maximum. Cependant dans le cas général cet algorithme, comme toute
heuristique, ne produit qu'un résultat approché, c'est-à-dire que
l'algorithme s'arrête à partir du moment où la modularité ne peut
plus augmenter. Or la fonction de recouvrement montre que certains
sommets seraient incités à changer de communauté et que ce changement
pourrait améliorer où inversement abaisser le résultat obtenu. Autrement
dit le résultat obtenu n'est pas stable. La recherche de communautés
stables a fait l'objet de peu de publications, les auteurs se satisfaisant
en général des résultats de l'algorithme qu'ils mettent en oeuvre
et les comparant aux résultats d'autres auteurs. Cependant la recherche
de stabilité d'un réseau en termes de théorie des jeux a fait l'objet
de nombreuses publications comme par exemple \cite{Nisan2007}. Selon
cette approche la recherche de stabilité de n agents qui choisissent
des stratégies à partir de fonctions de satisfaction suppose l'existence
d'un équilibre de Nash. Appliquée à la détection de communauté le
problème consiste à trouver les conditions d'existence d'un équilibre
de Nash tel qu'aucun sommet ne souhaite au final quitter la communauté
à laquelle il a été affecté. A notre connaissance l'équilibre de Nash
n'a été que peu utilisé pour détecter des communautés. \cite{RNarayanam2012}
l'appliquent sur des graphes unipartis. Ils utilisent la connectivité
des sommets pour atteindre un équilibre de Nash sans mesurer la modularité
du résultat. \cite{Chen2011} se focalisent aussi
sur les graphes unipartis et les communautés sont recouvrantes et
la recherche de l'équilibre de Nash est l'unique principe directeur.
Mais les résultats expérimentaux ne nous semblent pas en rapport avec
ce que à ce qui a pu être publié sur le sujet. Dans le présent article
nous nous distinguons de ces auteurs en recherchant d'abord une solution
approchée robuste à l'aide de l'algorithme de Louvain, puis effectuons
des réaffectations pour converger vers un équilibre local qui est
prouvé être un équilibre de Nash. De plus nous pouvons appliquer cette
approche sur les trois types de graphes.
\section{Détection de communautés partitionnées et recouvrement}
\subsection{Modularité}

Nous avons montré dans \cite{Crampes2013} que la recherche de communauté
partitionnées dans les graphes unipartis, bi-partis et orientés pouvait
se ramener à la recherche de communautés dans les graphes uni-partis.
En conséquence nous établissons les définitions et démontrons les
propriétés pour les graphes uni-partis puis lors des expérimentations
nous appliquerons les résultats théoriques à des cas de graphes bi-partis.
Le consensus parmi les auteurs pour la détection de communautés consiste
à rechercher une solution approchée qui maximise la modularité selon
\cite{Newman2006}. Cet indice mesure la qualité de la partition d'un
graphe. Formellement, étant donné un graphe uniparti $G=(N,E)$ représenté
par sa matrice d'adjacence $A$ , la modularité $Q$ d'une partition
de graphe est définie:
\begin{eqnarray}
Q=\frac{1}{2m}\sum_{i,j}\left[A{}_{ij}-\frac{k_{i}k_{j}}{2m}\right]\delta(c_{i},c_{j})
\end{eqnarray}
où $A{}_{ij}$represente le poids de la liaison entre $i$ et $j$,
$k_{i}=\sum_{j}A{}_{ij}$ est la somme des poids des arcs attaché
au sommet $i$, $c_{i}$ est la communauté à laquelle appartient le
sommet $i$, la fonction de Kronecker $\delta(u,v)$ est égale à $1$
si $u=v$ et à $0$ sinon, enfin $m=1/2\sum_{ij}A{}_{ij}$. Pour l'instant
nous ne considérons que des graphes binaires et dans ce cas les poids
$A{}_{ij}$ prennent les valeurs 1 ou 0 selon que la liaison existe
ou n'existe pas. L'interprétation de cette formule est la suivante
: la modularité est la somme pondérée pour toutes les communautés
de la différence entre les liaisons observées à l'intérieur de la
communauté (terme $A{}_{ij})$ et la probabilité de ces liaisons (terme
$\frac{k_{i}k_{j}}{2m}$ dont le numérateur est le produit des marges
correspondant à la cellule i,j). L'application de cette fonction par
de nombreux algorithmes donnent de bons résultats. Par exemple elle
permet de retrouver des communautés sur des graphes construits ad-hoc.
Cependant il a été démontré dans \cite{Fortunato2009} que cette fonction
a tendance à fusionner les petites communautés et ainsi à masquer
une certaine granularité. Pour les graphes bipartis (et les graphes orientés qui peuvent se
ramener à des graphes bi-partis) d'autres formulations ont été proposées.
En particulier la formulation dans \cite{Barber2009} semble faire
consensus. Peu différente de celle de Newman, nos travaux récents
montre qu'elle pourrait aussi s'appliquer aux graphes unipartis. Cependant
nous utiliserons la formulation de Newman quel que soit le type de
graphe pour rester conforme à nos travaux antérieurs publiés afin
d'unifier les trois types de graphes.

\subsection{Détection de communautés partitionnées}
Parmi les très nombreux algorithmes proposés dans la littérature pour
détecter des communautés partitionnées l'algorithme de Louvain \cite{Blondel2008}
est remarquable pour son efficacité et la qualité de ses résultats.
Il utilise une méthode hiérarchique pour constituer les communautés
en trouvant à chaque étape l'optimum de modularité pour chaque sommet.
A chaque étape chaque communauté constitue un nouveau sommet qui représente
tous ses sommets qui sont eux ôtés du nouveau graphe. Ainsi le graphe
se réduit progressivement pour obtenir la modularité optimum.
\subsection{Détection et analyse de communautés recouvrantes\label{sec4}}
Plusieurs approches sont possibles pour identifier le recouvrement
entre communautés. Dans la littérature en général le recouvrement
fait intrinsèquement parti de la recherche de communautés, les sommets
étant affectés par l'algorithme à plusieurs communautés. C'est par
exemple le cas dans \cite{Palla2005a} et dans\cite{Chen2011}.
Dans notre cas, nous identifions d'abord les communautés partitionnées,
puis une fonction de recouvrement qui permet de faire un premier bilan
de l'instabilité de l'affectation. Nous obtenons ainsi à la fois des
communautés partitionnées et leur recouvrement. Plusieurs fonctions de recouvrement sont possibles. Certaines ne font
apparaître que l'existence d'un recouvrement. D'autres comme celle
que nous présentons ici permettent de mesurer jusqu'à quel point deux
communautés se chevauchent. Notre fonction que nous appelons la 'légitimité'
présente aussi le grand intérêt de détailler pour chaque sommet son
degré d'appartenance à différentes communautés. Ainsi pour chaque sommet $u_{i}$, son degré d'appartenance (ou de
légitimité d'appartenance) à la communauté $c$ est mesuré par son
nombre de liens avec les autres sommets de la communauté divisé par
la taille de cette communauté.
\begin{eqnarray}
L(u_{i}\in c)=\frac{\sum{}_{j}A_{ij}\delta(c_{j})}{|\{v\in c\}|}
\end{eqnarray}

Cette fonction peut s'interpréter de la manière suivante: un sommet
est attiré par une communauté d'autant plus qu'il a un nombre relatif
élevé de relations avec cette communauté indépendamment de la taille.
Elle présente l'avantage d'être simple et intuitive. Par exemple dans
le benchmark SW présenté dans la section 6 on peut observer que $c_{1}$
contient 7 évènements, $c_{2}$ 5 évènements et $c_{3}$ 2 évènements.
On obtient pour $w_{1}$ une légitimité de $\frac{2}{7}$ en direction
de $c_{1}$, $\frac{1}{5}$ en direction de $c_{2}$ et $\frac{1}{2}$
en direction de $c_{3}$. Bien que simple, cette fonction présente
des propriétés intéressantes que nous ne pouvons pas démontrer dans
les limites de cet article. En particulier il s'agit là d'une fonction
d'appartenance floue sur laquelle il est possible d'appliquer une
$\alpha-coupe$ pour mieux observer les recouvrements et surtout les
sommets déviants (mal classés).

\section{Réaffectation}

\subsection{Fonction de réaffectation RM}
La fonction de légitimité montre comme dans l'exemple précédent que
certains sommets sont à cheval sur certaines communautés et qu'en
conséquence les communautés identifiées peuvent être instables. L'algorithme
de Louvain, comme tous les autres algorithmes donne une solution approchée.
Pour rechercher une meilleure stabilité de chaque communauté, nous
introduisons une fonction de réaffectation qui vise à améliorer la
modulartité globale. Réaffecter le sommet $w$ de $C_{1}$ à $C_{2}$ accroit ou décroit
la modularité. Nous définissons ce changement comme la mesure de réaffectation
de modularité. 
Soit $w$ un sommet. Si $w$ est retiré de $C_{1}$ et réaffecté à 
$C_{2}$, alors nous pouvons définir $RM_{w:C_{1}\rightarrow C_{2}}$
= $Q_{w\in C_{2}}$- $Q_{w\in C_{1}}$ où $Q_{w\in C_{2}}$ et $Q_{w\in C_{1}}$
sont les modularités développées ci-après. Soit $l_{w|i}=l_{w,w'|w'\in C_{i}}$ le nombre d'arêtes entre un sommet
$w$ et tous les autres sommets $w'$ tels que $w'\in C_{i}$, soit
$d_{w}$ le degré de $w$, \textbar{}$e_{i}|$ le nombre d'arêtes
dans $C_{i}$ et $d_{C_{i}}$= $\sum$ $d_{u|u\in c_{i}}$. Nous considérons
que le sommet $w$ appartenant à $C_{1}$ est retiré de cette communauté
et ensuite affecté à la communauté $C_{2}$, 


alors $RM_{w:C_{1}\rightarrow C_{2}}=Q_{w\in C_{2}}$-$Q_{w\in C_{1}}$
= $[\frac{1}{m}(|e_{1}|-l_{w|1})+\frac{1}{m}(|e_{2}|+l_{w|2})-(\frac{(d_{C_{1}}-d_{w})^{2}}{(2m)^{2}}+\frac{(d_{C_{2}}+d_{w})^{2}}{(2m)^{2}})]-[\frac{1}{m}|e_{1}|-\frac{(d_{C_{1}})\text{\texttwosuperior}}{(2m)^{2}}+\frac{1}{m}|e_{2}|-(\frac{(d_{C_{2}})\text{\texttwosuperior}}{(2m)^{2}})]$. 

Et après simplification, 
\begin{eqnarray}
RM_{w:C_{1}\rightarrow C_{2}}=\frac{1}{m}(l_{w|2}-l_{w|1})-\frac{1}{2m{}^{2}}[d_{w}^{2}+d_{w}(d_{C_{2}}-d_{C_{1}})]\label{eq:11}
\end{eqnarray}
 

Cette équation peut être partiellement validée par quelques opérations
simples. Quand on retire $w$ à $C_{1}$ puis quand on le réaffecte
à $C_{1}$ la modularité devrait retrouver sa valeur initiales c.à.d.
$RM_{w:C_{1}\rightarrow C_{1}}=0$. En effet en considérant que $C_{2}$
est dans ce cas là $C_{1}$ sans la présence de $w$, nous obtenons
$d_{C_{2}}=d_{C_{1}}-d_{w}$, et en remplaçant $d_{C_{2}}$ dans l'équation
\ref{eq:11} par sa valeur, nous obtenons bien $RM_{w:C_{1}\rightarrow C_{1}}=0$. 
On trouve une formulation voisine obtenue à partir d'un calcul différent dans \cite{Wu2012a}. 

\subsection{Effet de la réaffectation sur les autres sommets}
Nous présentons ci-dessous une simplification du calcul de la réaffectation
pour les autres sommets. Nous supposons, qu'une première passe de calcul des réaffectation
a été effectuée, et à la suite de cela, $w$ un sommet du graphe a
été déplacé de $C_{1}$ à $C_{2}$, suite à une mesure de réaffectation
positive. Soit $z$ un autre sommet du graphe. On peut voir la variation
de la valeur de réaffectation pour ce sommet après le déplacement
de w. Le calcul de la différence de la mesure de réaffectation pour le sommet
$z$ de l'étape précédente à l'étape actuelle se calcule comme suit : 
On cherche $RM_{z:C_{from}\rightarrow C_{to}}^{1}-RM_{z:C_{from}\rightarrow C_{to}}^{0}$
où $RM_{z:C_{from}\rightarrow C_{to}}^{0}$
est la mesure de réaffectation de z avant le déplacement de w et $RM_{z:C_{from}\rightarrow C_{to}}^{1}$
sa mesure après. Selon les cas de $C_{from}$ et $C_{to}$ on trouve
les résultats qui suivent. Soit $\triangle R_{z}=\{w,z\}-\frac{1}{(2m)}d_{z}d_{w}$,
dans lequel $\{w,z\}$ représente le lien entre $w$ et $z$. S'il
n'y a pas de lien cette valeur est nulle. Dans le tableau ci dessous, nous montrons les différentes valeurs
de la correction de réaffectation pour $z$. 

\begin{tabular}{|c|c|c|c|c|}
\hline 
to\textbackslash{}from & $C_{1}$ & $C_{2}$ & $C_{3}$ & $C_{4}$\tabularnewline
\hline 
\hline 
$C_{1}$ & 0 & -2$m\triangle R_{z}$ & -$m\triangle R_{z}$ & -$m\triangle R_{z}$\tabularnewline
\hline 
$C_{2}$ & 2$m\triangle R_{z}$ & 0 & $m\triangle R_{z}$ & $m\triangle R_{z}$\tabularnewline
\hline 
$C_{3}$ & $m\triangle R_{z}$ & -$m\triangle R_{z}$ & 0 & 0\tabularnewline
\hline 
$C_{4}$ & $m\triangle R_{z}$ & -$m\triangle R_{z}$ & 0 & 0\tabularnewline
\hline 
\end{tabular}

On peut relever les propriétés suivantes : 

$m(RM_{z:C_{1}\rightarrow C_{2}}^{1}-RM_{z:C_{1}\rightarrow C_{2}}^{0})=$$m(RM_{z:C_{1}\rightarrow C_{3}}^{1}-RM_{z:C_{1}\rightarrow C_{3}}^{0})+m(RM_{z:C_{3}\rightarrow C_{2}}^{1}-RM_{z:C_{3}\rightarrow C_{2}}^{0})$
= $2\triangle R_{z}$ 

et $m(RM_{z:C_{2}\rightarrow C_{1}}^{1}-RM_{z:C_{2}\rightarrow C_{1}}^{0})=m(RM_{z:C_{2}\rightarrow C_{3}}^{1}-RM_{z:C_{2}\rightarrow C_{3}}^{0})+m(RM_{z:C_{3}\rightarrow C_{1}}^{1}-RM_{z:C_{3}\rightarrow C_{1}}^{0})$
= $-2\triangle R_{z}$.

Ce tableau permet de simplifier les calculs de réaffectation. Il est
aussi un moyen intéressant d'étudier l'impact sémantique 
de la réaffectation sur les autres sommets.
Sans détailler les conclusions dans les limites de cet article, on
note qu'un sommet a tendance à suivre un sommet voisin avec lequel il a une liaison, ou bien à
quitter une communauté dans laquelle arrive un sommet avec lequel
il n'a pas de liaison.
\vspace{-0.5 cm}

\section{Réaffectation et Équilibre de Nash (EN)}
\vspace{-0.2 cm}
Nous montrons dans cette section que l'application de la réaffectation
conduit à un équilibre de Nash, une situation où aucun sommet n'a
intérêt à quitter la communauté à laquelle il a été finalement affecté. 
Nous pouvons ramener le problème de la réaffectation des sommets à
un problème de théorie des jeux. En effet les n sommets peuvent être
considérés comme n agents qui cherchent à opitimiser leurs gains en
jouant des stratégies. 
Soit un jeu à un ensemble fini N de joueurs, chaque joueur $i$ ayant
accès à un ensemble fini $S_{i}$ de stratégies : $S_{i}=({s_{i_{1}},s_{i_{2}}...,s_{i,m_{i}}})$.
Une stratégie est un déplacement accessible à un joueur pour lequel
il est en droit d'attendre un gain. Ce gain dépend des stratégies
des autres joueurs. Formellement un jeu $G=(S, f)$ où $S=S_{1}\times S_{2}...\times S_{n}$
est l'ensemble des profils de stratégies. Un profil de stratégie (ou
vecteur stratégie) $s$$\in S$ est la combinaison des stratégies
de tous les joueurs à un moment du jeu, chaque joueur appliquant une
seule stratégie. A chaque profil de stratégies correspond une fonction
de gain $f_{i}:S\longrightarrow\mathbb{R}$ pour le joueur i et $f=({f_{1},f_{2},...,f_{i},...,f_{n}})$.
Il est important de noter que le gain du joueur $i$ dépend à un instant
donné de la stratégie de l'ensemble des joueurs. 
Un profil de stratégie $s^{*}$ est un équilibre de Nash (EN) si aucun
joueur n'a intérêt à changer de stratégie étant donné la stratégie des autres joueurs. 
Formellement : Un profil de stratégie $s^{*}$ est un équilibre de Nash
(EN) si pour tous les joueurs $i$ et pour toute stratégie alternative
$s'_{i}\in S_{i}$, $s'_{i}\neq s_{i}$ , $f_{i}(s_{i},s_{-i})\geq f_{i}(s_{i}',s_{-i})$
où $s_{-i}$ est un vecteur stratégie duquel a été retirée la stratégie
de l'agent $i$. 
Un jeu à stratégies pures est un jeu pour lequel les joueurs jouent
de manière déterministe; un jeu à stratégies mixtes est un jeu dans
lesquel les joueurs font intervenir une fonction de probabilité pour
décider du coup à jouer. 
Le théorème de Nash affirme que dans un jeu à n joueurs (n fini) à
stratégies mixtes il existe au moins un équilibre, c'est-à-dire un
profil $s^{*}$ tel que connaissant les stratégies des autres joueurs,
aucun joueur n'a intérêt à changer de stratégie. Ce théorème vrai
pour les stratégies mixtes l'est aussi pour les stratégies pures qui
sont des cas particuliers des stratégies mixtes pour lesquels la probabilité
est égale à 1.

Deux remarques s'imposent. D'une part cet équilibre peut ne pas être
unique. D'autre part rien n'indique comment atteindre cet équilibre;
dans la plupart des cas l'obtention d'un EN est un problème NPComplet
(en fait la problématique de la NPComplétude est plus... complexe;
le lecteur intéressé en trouvera une présentation dans Papadimitriou
\cite{Nisan2007}). 
Une réponse partielle à la seconde remarque peut être trouvée si on
peut définir une fonction potentiel qui permet alors, en recherchant
des optima locaux pour les agents d'atteindre un optimum global. Si
de plus la recherche d'un optimum local pour un agent peut se faire
en un temps polynomial, alors le problème rentre dans la classe des
problèmes PLSComplet (Polynomial Local Search) (Eva Tardos \& Tom
Wexler\cite{Nisan2007}). 
Pour tout jeu fini, une fonction potentiel exacte $\Phi$ est une
fonction qui fait correspondre à chaque vecteur stratégie s une valeur
réelle avec les conditions suivantes : $\Phi(s')-\Phi(s)=f_{i}(s')-f_{i}(s)$, 
où $s=(s_{i},s_{-i})$ et $s'=(s'_{i},s_{-i})$. Cette fonction s'interprète de la manière suivante: si le joueur i
passe unilatéralement d'une stratégie $s_{i}$ à une stratégie $s'_{i}$
avec un gain $f_{i}(s')-f_{i}(s)$ la fonction potentiel évolue d'autant.
Ainsi armé d'une fonction potentiel il est possible de trouver sûrement
un équilibre de Nash en convergeant à partir de la recherche d'optima
locaux pour chaque joueur. 

Dans le cas de la détection des communautés les agents sont les n
sommets et les stratégies sont les communautés auxquelles les sommets
souhaitent être rattachés. Pour trouver sûrement de manière convergente
un équilibre de Nash, c'est-à-dire une partition des communautés qui
satisfasse tout le monde, il nous faut construire une fonction potentiel
$\Phi(s)$. La fonction $RM_{w:C_{1}\rightarrow C_{2}}$ qui représente le gain
attendu par le sommet $w$ quand il est réaffecté est en fait le gain
de modularité pour toutes les communautés. Autrement dit $\Phi(s)=f_{i}(s)$. 
Ce choix de gain local assure donc que l'algorithme local de réaffectation
des sommets présenté maintenant converge bien vers
un équilibre de Nash. 

En pratique, partant d'une partition déjà établie, nous calculons la valeur de $RM$ pour tous les sommets vis à vis de toutes les communautés. Tous les sommets qui ont une valeur positive sont instables. Nous choisissons la valeur la plus élevée et réaffectons le sommet dans la communauté désirée. Les $RM$ des sommets des deux communautés concernées (départ et arrivé) sont recalculées pour la réaffectation suivante. L'algorithme s'arrête quand tous les $RM$ sont nuls ou négatifs. L'équilibre est atteint. Ainsi à partir d'un quelconque algorithme de partitionnement il est possible de vérifier la stabilité et réaffecter des sommets jusqu'à atteindre un équilibre de Nash, pour lequel toutes les affectations sont acceptées.  Le calcul
 est polynomial, l'algorithme de convergence vers l'équilibre de Nash est donc PLSComplet. A l'équilibre
tous les sommets ont une valeur de $RM$ négative. 

\vspace{-0.5 cm}
\section{Experimentation\label{sec5}}
\vspace{-0.2 cm}
Dans la mesure où nous avons montré dans \cite{Crampes2013} qu'il
est possible de traiter indifféremment des graphes unipartis, bipartis
et orientés, nous avons effectué deux types d'expérimentation. D'une
part nous avons appliqué notre algorithme de réaffectation à dà des graphes unipartis traditionnels
 tels que le 'karate club' ou les 'dauphins'
pour comparer nos résultats à d'autres auteurs. D'autre part nous
l'avons appliqué à  la détection de communauté dans des graphes bipartis.
Nous laissons le lecteur prendre connaissance de ces graphes dans
notre article ci-dessus, ou pour les dauphins dans d'autres articles
cités en référence.
\vspace{-0.5 cm}

\subsection{Graphes unipartis}
\vspace{-0.2 cm}
Louvain appliqué seul au graphe 'karaté' produit quatre communautés.
Notre mesure de réaffectation ne montre aucune instabilité 
puisqu'il n'y a pas de valeurs de $RM$ positives. La modularité
mesurée est supérieure à celle de tous les autres algorithmes que
nous connaissons. En particulier elle est nettement meilleure (0,470)
que celle déclarée chez les auteurs qui utilisent aussi une approche
locale pour obtenir un équilibre de Nash, en particulier chez \cite{RNarayanam2012}
qui produisent 3 communautés ainsi que chez  \cite{Chen2011} qui détectent aussi 3 communautés.
En ce qui concerne les dauphins, Louvain seul produit 4 communautés
avec une modularité de 0,48 et notre mesure de réaffectation montre
l'instabilité de trois sommets. Après réaffectation de ces sommets,
à l'équilibre, la modularité monte à 0,51. Les autres auteurs affichent
une modularité résultante moindre sauf \cite{RNarayanam2012} pour
lequel l'absence de données ne nous a pas permis de vérifier les résultats.
Sans analyser plus loin, il nous paraît que notre approche améliore
de manière intéressante la méthode de Louvain, surtout en lui offrant
un moyen polynomial de vérifier la stabilité. Comparé aux autres méthodes
qui utilisent l'équilibre de Nash, nous faisons mieux, ou, dans certains
cas presque aussi bien (pour peu qu'il soit possible de vérifier les
résultats de ces auteurs). 
\vspace{-0.5 cm}

\subsection{Graphes bipartis}
\begin{figure}
\includegraphics[width=11cm,height=3.9cm]{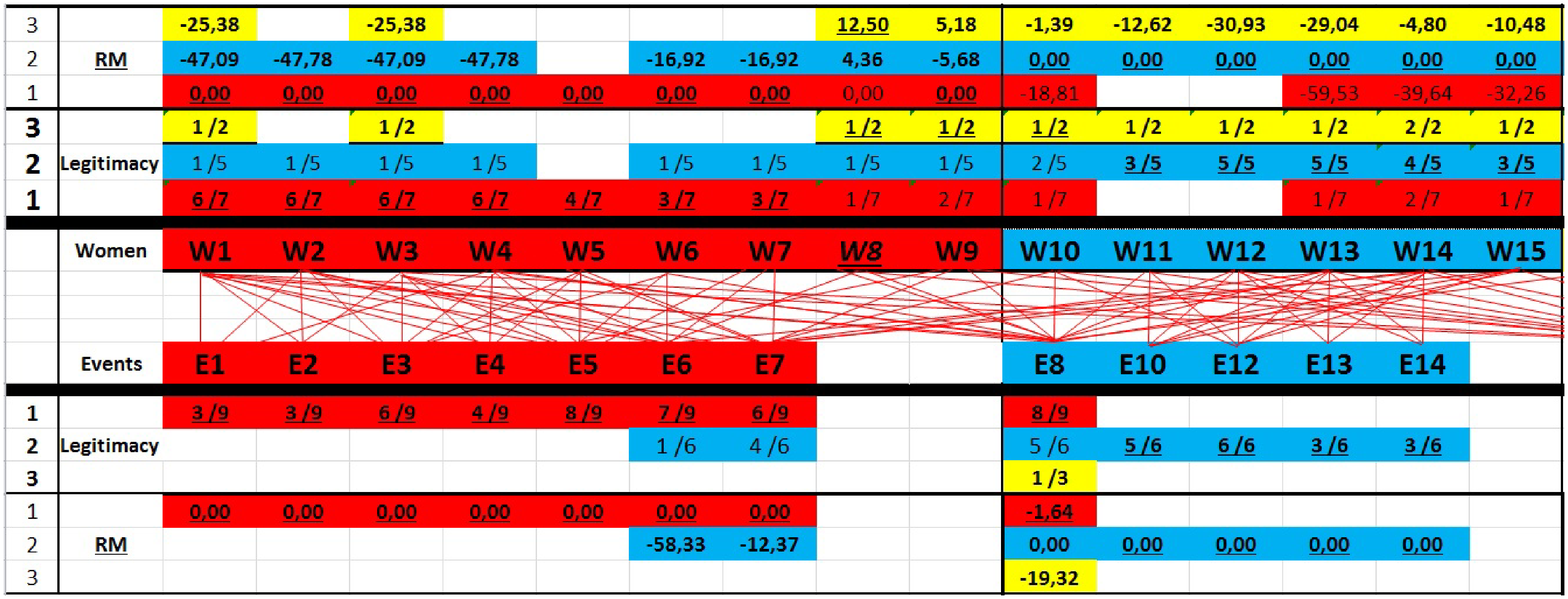}
\vspace{-0.2 cm}
\caption{Communautés et mesures de modularités pour : Women Events}
\label{WE}
\end{figure}

\vspace{-0.2 cm}
\textbf{Jeu de données 'Southern Women.' \label{sub:Southern-Women}}
Des résultats partiels ont déjà été présentés dans \cite{Crampes2013} sans réaffectation. 
Nous le complétons ici avec le recouvrement et l'équilibre de Nash.
Dans la figure \ref{WE}, le graphe biparti est montré comme un graphe
à deux couches au milieu avec les dames sur la couche du dessus et
les événements sur la couche du dessous. Les arêtes entre dames et
événements représentent leurs participations à ces événements. Trois
communautés regroupant des dames et des événements ont été détectées
et montrées dans cette figure en couleur rouge, bleu et jaune. Ce
résultat est plus précis que la majorité des résultats présentés dans
\cite{Freeman}. La figure montre les recouvrements en utilisant deux
mesures : la légitimité et la mesure de réaffectation de modularité
(RM) pour les dames et les évènements. Pour les dames ces deux mesures
sont représentées en fonction des communautés partitionnées au dessus
des dames. Les meilleures valeurs maximales pour la légitimité et RM sont soulignées.
Seule une dame (W8) - en italique - présente des valeurs de légitimité
et de RM supérieures pour des communautés autres que celles auxquelles
elle a été affectée. Notre méthode met ainsi en relief des litiges
d\textquoteright{}affectation qu\textquoteright{}aucun autre auteur
n\textquoteright{}avait fait apparaitre. La mesure de RM montre avec
plus d'accuité cette tendance à être affecté de façon préférentielle
à une autre communauté. Concrètement deux individus
(W8 et W9) sont instables. Après réaffectation l'ensemble des trois
communautés est stable. La modularité passe de 0,309 à 0,325. Il est
intéressant de noter que les deux individus réaffectés le sont dans
la troisième communauté qui se trouve renforcée alors qu'aucun autre
auteur ne l'avait identifiée.

\vspace{-0.2 cm}
\textbf{Compte Facebook} Cette expérimentation déjà présentée en partie dans \cite{Crampes2013}
mais sans réaffectation, traite d'un compte facebook d'un étudiant dont nous avons extrait
les photos et les tags associés aux photos. Le jeu de données comporte
644 photos pour 274 tags différents. Après curetage et un premier
passage par Louvain 16 communautés de 2 personnes ou plus sont identifiées.
Elles sont peu recouvrantes. La mesure RM nous montre qu'un seul
sommet sur plus de 250 personnes est instable. Ceci indique une pertinence
particulièrement élevée des communautés partitionnées d'origine. La
mesure RM appliquée aux photos montre également que 15 photos souhaitent
être affectées à des communautés différentes de celles trouvées par
le partitionnement. Ceci indique que dès le début le jeu de données
révèle peu d'ambiguité. Ceci peut se comprendre étant donné le fait
que les photos supposent la présence physique des individus dans un
temps et un lieu précis. Elles témoignent assez bien de périodes de
la vie de celui qui les possède.

\vspace{-0.5 cm}
\section{Perspectives et Conclusion}
\vspace{-0.2 cm}
Nous avons appliqué une méthode de détection de communautés dans des
graphes unipartis et bipartis. Notre première contribution a consisté
à faire apparaitre les recouvrements. La deuxième contribution consiste
à faire apparaître l'instabilité des résultats produits par l'algorithme
de détection malgré leur qualité. Pour trouver une solution stable
nous avons fait appel à une fonction de réaffectation et nous avons
démontré que cette solution nous permettait d'atteindre un équilibre
de Nash en temps polynomial. Plus généralement, presque toutes les
méthodes de détection de communautés produisent des situations d'affectation
instables. Notre contribution permet de compléter tout algorithme
de détection de communautés pour trouver en temps polynomial un équilibre
stable pour lequel la modularité est optimale. On notera pour ce faire
le caractère 'social' de tous les sommets qui identifient leur intérêt
à l'intérêt collectif. Ce point sera approfondi dans l'une des suites
à nos travaux. Nos recherches futures porteront sur l'amélioration des
méthodes de réaffectation, Une limite à la méthode présentée est le fait 
qu'elle prend pour contrainte le nombre de communautés trouvé par 
l'algorithme initial de détection. Nous étudierons comme supprimer 
ou faire apparaître de nouvelles communautés. Le but
est d'étudier comment des politiques de rajout ou de suppression de communautés
peuvent entrainer des instabilités et des insatisfactions. Plus globalement
si la création d'un graphe initial ainsi que la détection de communautés
ont été largement étudiées, il existe peu de travaux sur l'application
de contraintes externes et sur la dynamique d'adaptation des communautés.
Ces modèles trouveraient leur application à des problèmes d'ajout ou à l'inverse de
 suppression
d'unités de production, de services, etc. La réalité suppose de disposer
de graphes multipartis ce qui explique notre approche qui mèle dans
un premier temps graphes unipartis et graphes bipartis. De même il
faudra considérer des fonctions potentiels plus égoïstes qui laissent
aussi la place à des agents moins altruistes.

\vspace{-0.5 cm}
\bibliography{communities}

\end{document}